\documentclass{article}
\usepackage[margin=1in]{geometry} 
\usepackage[table]{xcolor}
\definecolor{Gr}{gray}{0.9}
\usepackage{bm}
\usepackage{graphicx,stackengine}
\usepackage[percent]{overpic}
\usepackage{amsfonts}
\usepackage{hyperref,rotating}
\hypersetup{
    colorlinks=true,
    linkcolor=blue,
    filecolor=magenta,      
    urlcolor=cyan
    }
\usepackage{multirow}
\usepackage{amsmath}
\usepackage{booktabs}
\usepackage{tabularx} 
\usepackage{enumitem}
\usepackage[numbers, square,comma]{natbib}

\begin{document}

\title{Performance Assessment Strategies for Language Model Applications in Healthcare}

\author{Victor Garcia\footnotemark[1] \footnotemark[2] , Mariia Sidulova\footnotemark[2] \footnotemark[3] , Aldo Badano}
\date{\small\textit{Office of Science and Engineering Laboratories, Center for Devices and Radiological Health,
U.S. Food and Drug Administration, Silver Spring, MD, 20993, USA}}

\footnotetext[1]{Corresponding author. Email address: victor.garcia@fda.hhs.gov}
\footnotetext[2]{Victor Garcia and Mariia Sidulova contributed equally to this work.}
\footnotetext[3]{This author is currently affiliated with Medtronic.}
\footnotetext[4]{This article is published in \textbf{Artificial Intelligence in the Life Sciences} and is available at: \url{https://doi.org/10.1016/j.ailsci.2026.100162}}

\maketitle

\begin{abstract}
Language models (LMs) represent an emerging paradigm within artificial intelligence, with applications throughout the medical enterprise. A comprehensive understanding of the clinical task and awareness of the variability in performance when implemented in actual clinical environments lays the foundation for the LM application assessment. Presently, a prevalent method for evaluating the performance of these generative models relies on quantitative benchmarks. Such benchmarks have limitations and may suffer from train-to-the-test overfitting, optimizing performance for a specified test set at the cost of generalizability across other tasks and data distributions. Evaluation strategies leveraging human expertise and utilizing cost-effective computational models as evaluators are gaining interest. We discuss current state-of-the-art methodologies for assessing the performance of LM applications in healthcare and medical devices.
\end{abstract}

\vspace{10pt}

\section{Introduction}
Language model (LM) applications for text, images, audio, or code are poised to play a significant and growing role in healthcare\cite{Shanmugan2025}. The recent advances in generative model capabilities with transformer-based architectures have expanded the scope of artificial intelligence's (AI's) potential impact in healthcare. Recent reports have documented the potential of LM technologies for tasks such as clinical records documentation, clinical encounter summarization, and patient management, as well as in imaging applications for synthetic data generation\cite{Sizikova2024}, image enhancement and reconstruction\cite{Jung2024}, and segmentation of anatomical structures or lesions\cite{ZHANG2024108238}.  

A burgeoning area of LM applications has been in text-based analysis and generation. VanVeen et al. have applied language models to clinical summarization tasks related to radiology reports, patient questions, and problem lists generation and found models to be either non-inferior (45\%) or preferred (36\%) compared to medical expert summaries, without an increase in harm\cite{van2024adapted}. Osborne et al. have compared models against clinicians in the generation of discharge summaries from the clinical record and found that model-generated summaries are preferred over clinician-generated summaries in dimensions such as quality and readability/conciseness while remaining comparable for factuality and completeness\cite{osborne2025towards}. Beyond summarization, multi-modal models have demonstrated potential for the interpretation of medical images\cite{zhong2025vision, tanno2025collaboration, pellegrini2025radialog}. As part of the assessment of these open-ended outputs, Zhong et al. utilize metrics such as area under the receiver operating characteristic (AUROC), sensitivity, specificity, and others for the multi-classification assessment of whether an abnormality was correctly detected, while separately assessing report quality with text comparison metrics such as BLEU, METEOR, CIDEr, and others\cite{zhong2025vision}. Tu et al. and Nori et al. have developed systems with AI agents to assess a model's ability to diagnose medical conditions based on clinical vignettes while imitating the multi-turn conversations that occur between the provider and patient\cite{tu2025towards, nori2025sequential}. 

These few examples highlight how each application of AI may benefit from a different assessment method. Benchmark datasets can facilitate head-to-head comparison of different models on the same data while utilizing natural language metrics. Human experts can be utilized to provide both reference standards for proportion-based metrics and to perform preference studies, and models as evaluators may support the human evaluation. Independent of the use case, a rigorous evaluation of LM applications in medicine is an essential step in the development and continuous monitoring of clinical solutions that carry direct risks to patients in the sensitive context of clinical care\cite{Mesko2023}. At the same time, evaluation strategies need to be practical and scalable to ensure timely technological design cycles. To better understand the trade-off between rigor, practicality, and scalability, we propose a high-level categorization of evaluation strategies with the three categories of benchmark evaluation, human evaluation, and model-based evaluation with a discussion of the advantages and disadvantages of each.

\section{Benchmark Evaluation}
Benchmark evaluation is the evaluation of models on established testing datasets with predetermined metrics. This approach is as popular as it is practical, allowing for head-to-head model comparisons on the same data at scale. Though this definition specifies the use of an established testing dataset as a benchmark, non-benchmark datasets may also be repurposed to create novel datasets\cite{Blagec2023-JBI}, but additional considerations may be needed for these secondary use applications to avoid data leakage from model development\cite{samala2021risks}. 

\subsection{Description}
Benchmark evaluations typically involve tasks such as short-answer responses or multiple-choice questions. Multiple-choice question benchmarks may utilize proportion-based metrics, such as in the Massive Multitask Language Understanding (MMLU) dataset\cite{hendrycks2020measuring}. MMLU consists of approximately 16,000 multiple-choice questions spanning 57 subjects, including mathematics, history, and computer science, designed to evaluate a model's multitask accuracy. Other benchmarks may include the evaluation of short-answer responses, such as in General Language Understanding Evaluation (GLUE)\cite{wang2018glue}. GLUE is comprised of nine diverse language understanding tasks designed to evaluate models' abilities in areas like sentiment analysis, sentence similarity, and textual entailment, and includes correlation metrics, accuracy, and F1-score.  

Medical benchmarks for LM models can be categorized based on modality, such as text, image, and multimodal data\cite{yan2024large}. These benchmarks can be further classified according to their focus areas, including electronic health records, doctor-patient dialogues, medical question-answering, and medical image captioning. A prominent text benchmark is the MedQA dataset \cite{jin2021disease}, which consists of multiple-choice questions derived from medical board exams, including the United States Medical Licensing Examination (USMLE), and is commonly used to assess models' understanding and application of medical knowledge in clinical scenarios. As part of MedQA, models are tasked to identify relevant information from a medical corpus to inform the model’s inference, with OpenAI models achieving accuracies above 95\%\cite{medqabenchmark, bedi2025fidelity, scienceandmedicine}. ReXVQA is a medical image question-answer benchmark that assesses a model’s interpretation of chest X-rays, and Google’s MedGemma 4B has been shown to outperform radiology residents with an accuracy of 84\% compared to the closest performing reader’s accuracy of 77\%\cite{pal2025rexvqa}. Aggregate benchmarks have also been created that combine multiple smaller benchmark datasets of varying tasks, such as MedHELM, which is a benchmark suite comprised of 35 benchmarks encompassing 121 medical tasks\cite{bedi2025medhelm, medhelmleaderboard}. 

Benchmarks are also developed for conversational and agent-based applications\cite{xiaolan2025evaluating}. To better replicate performance for conversational applications, HealthBench, SDBench, and MIMIC-CDM utilize multi-turn conversations to gather information rather than provide all information at one time through a clinical vignette. HealthBench consists of 5,000 multi-turn conversations between a healthcare provider and a model. In this benchmark, models are tasked to respond to the last user message, and the open-ended outputs are scored using a rubric applied by a model as evaluator (MAE)\cite{arora2025healthbench, simplevals}. In the work by Arora et al., OpenAI’s o3 has the average highest score of 0.60 compared to other tested models\cite{arora2025healthbench}. 

SDBench, or Sequential Diagnosis Benchmark, utilizes an AI agent to convert clinical vignettes from the New England Journal of Medicine’s Case Challenge series into multi-turn conversations only returning specific information related to the information requested by the model being assessed\cite{nori2025sequential}. As of this publication, SDBench is not yet publicly available. MIMIC-CDM provides a curated dataset of 2,400 patient cases from the Medical Information Mart for Intensive Care (MIMIC-IV)\cite{johnson2023mimic} where models must iteratively request and assess the provided clinical data to diagnoses one of four pathologies (appendicitis, pancreatitis, cholecystitis, and diverticulitis)\cite{hager2024evaluation, hager_mimic-iv-ext_nodate}. Benchmarks are also being developed to assess the tool calling capabilities of agent-based models. MedAgentBench contains 300 clinician-developed tasks that require models to utilize tools using Fast Healthcare Interoperability Resources (FHIR)-based API (application programming interface) calls, with Claude 3.5 Sonnet v3 having the highest overall success rate of 70\%\cite{jiang2025medagentbench}. 

\subsection{Advantages and Limitations}
The primary advantage of benchmarking is facilitating direct comparisons between models using identical datasets and metrics, promoting transparency and competition. Platforms such as Hugging Face \cite{huggingfaceleaderboards} provide leaderboards that display model rankings based on benchmark performance, helping researchers and practitioners select appropriate models for specific tasks.

Benchmarking faces several limitations including dataset limitations, overfitting, lack of clinical representativeness, and insufficient coverage of clinically relevant tasks. In addition, in the medical space, benchmarks have suffered from data quality issues and data homogeneity~\cite{Mincu2022}. First, benchmarks may not accurately reflect real-world complexities indicating low coverage of task variability, resulting in models that excel in tests but struggle with practical applications \cite{siska2024examining, huang2024planning}. For example, Yao et al. evaluated large language models' clinical skills using a framework inspired by medical education's Objective Structured Clinical Examinations (OSCEs) and found that models performing well on traditional multiple-choice question-answer benchmarks like MedQA encountered significant challenges when applied to more complex clinical scenarios \cite{yao2024medqa}. Similarly, Schmidgall et al. designed AgentClinic-MedQA, a benchmark to assess models that perform multi-run conversations with tool calling, and they show that a model’s performance on MedQA does not predict its performance on AgentClinic-MedQA\cite{schmidgall2024agentclinic}. These performance variations highlight how benchmark selection impacts our interpretation of a model's performance, which may not always reflect its expected workflow.

Data leakage presents another significant challenge. The difficulty in tracing models' training datasets makes it challenging to identify potential leakage. "Training to the test" or model overfitting \cite{montesinos2022overfitting} is when models are overly optimized for benchmark datasets at the expense of generalization capabilities. When benchmark data has been used during training, model performance on the benchmark may be artificially inflated \cite{sainz2023nlp}. Research by Xia et al. demonstrated that subtle changes in the HumanEval benchmark dataset led to an average performance drop of 39\% across 51 models, suggesting potential overfitting \cite{xia2024top}. Overfitting and training on the dataset can result in impressive benchmark scores that fail to translate to real-world effectiveness. 

\section{Human Evaluation}
Human evaluation refers to strategies that rely on human experts to establish the reference standard for a given task and to evaluate the output of a LM model\cite{kellerhuis2025expert, bertens2013use}. Human expert evaluation provides nuance and complexity of the medical decision-making, often involving subtle cues and contextual understanding that are difficult for automated metrics to capture fully. Human experts may help identify potential risks, biases, or errors in LM-enabled medical device outputs that could have serious consequences for patients. 

\subsection{Description}
Human evaluation has been applied to assess the performance of medical devices using LM approaches. For instance, Tu et al.\cite{Tu2024} described evaluation methods to assess Med-PaLM Multimodal (Med-PaLM M), a generalist biomedical AI model designed to handle diverse tasks and modalities (e.g., clinical text, medical imaging, genomics) with a single set of model weights. For tasks requiring clinical judgment, such as radiology report generation, human experts (clinicians) conducted qualitative assessments. The study involved 246 chest X-ray reports, where clinicians performed side-by-side comparisons of Med-PaLM M-generated reports from different model sizes (12B, 84B, and 562B) versus human radiologist-written reports. Experts rated the AI outputs for clinical accuracy, relevance, and utility, and indicated preferences in blinded evaluations. In a four-way comparison of the reports, the radiologist-written reports were ranked best in 37\% of cases, while Med-PaLM M (84B) ranked best in 26\% of cases with a mean total error rate of 0.59\cite{Tu2024}.

A multi-category evaluation rubric was also used in the assessment of Med-Gemini. Med-Gemini is a family of medical AI models created by fine-tuning Google's Gemini on diverse medical data (radiology, pathology, genomics) to perform clinical tasks like image classification, visual question answering, and report generation\cite{yang2024advancing}. Human evaluations used 7 radiologists who assessed chest x-ray cases and CT cases using a 6-category rubric comparing AI reports to original radiologist reports based on clinical accuracy and patient management impact. Results showed that 43\% of abnormal and 57\% of normal reports generated by Med-Gemini were rated as superior or similar to the original report on the MIMIC-CXR dataset, and 65\% and 96\% of abnormal and normal cases, respectively, on a chest x-ray dataset from India. For report generation from Head/Neck 3D CT volumes, Med-Gemini-generated reports were found to be clinically acceptable in 53\% of cases but only superior to the reference radiologist report for abnormal and normal cases in 15\% and 18\% of cases, respectively.

In Zhang et al.'s work, they leverage human evaluation to determine error rates, human preferences, and level of correctness in radiology-related tasks\cite{Zhang2024biomedgpt}. They describe BiomedGPT, a lightweight vision-language foundation model designed as a generalist AI for diverse biomedical tasks across multiple modalities including medical image classification, visual question answering, report generation, and clinical text understanding. One radiologist evaluated BiomedGPT across radiology tasks. For the generation of the 'findings' section of chest x-ray radiology reports, BiomedGPT-B, a 182 million parameter model,  showed an 8 \% critical error rate, as compared to the reported human error rate of 6\%, and a critical omission rate of 7\%. For the assessment of the 'impression' section, the radiology evaluator preferred human-written impressions over BiomedGPT-B summaries in 52\% of cases, with no significant difference in correctness and harm ratings. 

Finally, Reinforcement Learning from Human Feedback (RLHF)~\cite{NEURIPS2024_c1f66abb} represents a novel hybrid approach that transforms traditional expert evaluation from a passive assessment into an active training mechanism, where human judgments directly shape model behavior through iterative reinforcement learning cycles. In RLHF, human assessments of model outputs are systematically collected and used to train a reward model that guides further refinement of the model. This iterative approach helps align the model with human expectations and values, resulting in enhanced quality and usefulness of responses. In a performance comparison of supervised fine-tuning (SFT) alone against RLHF plus SFT, Rawal et al. demonstrated that in classification, question and answering, and summarization tasks, the addition of RLHF resulted in a 5\%, 10\%, and 15\% relative improvement in model performance, respectively\cite{rawal2024enhancing}. While not strictly a traditional evaluation method, RLHF represents a dynamic form of human expert evaluation that directly incorporates feedback into the training process. 

\subsection{Advantages and Limitations}
Human evaluation is a multi-faceted and adaptable approach for assessing a model. Experts can not only establish reference standards for discrete outputs, such as determining the presence of a medical condition within an imaging study, but they can provide a nuanced assessment of a model's output in the clinical context, such as if an error in a model's output would lead to a change in medical management. This evaluation approach can assess the appropriateness of open-ended model outputs including intermediate conversations and the model reasoning or tool calling steps of agents. Because humans are adaptable, the human evaluation remains central to model evaluations, especially within complex systems. 

Even though human experts possess extensive knowledge and experience, their judgments can still be influenced by cognitive biases, personal beliefs, and systemic factors, which may lead to inconsistencies, disparities in decision-making, or unintended biases in medical assessments. This necessitates methodological safeguards like blinding, randomization, and structured evaluation frameworks to maintain assessment objectivity. Expert evaluation is also, and most importantly, resource-intensive. It can be time-consuming and expensive, especially for statistically powered studies with large sample sizes and multiple experts, making them impractical for scaling to large datasets or to a large number of LM models. Though this may be mitigated through other methods, such as crowd-sourced expert annotations, understanding the quality of collected annotations and level of expertise of collectors becomes integral to the collection process. 

Human annotations are variable. Even for quantitative annotations created by qualified experts, inter-reader variability is considered when establishing reference standards. As a result, human evaluation strategies are affected by considerations regarding expert selection in terms of expertise and numbers of clinicians in expert panels. For subjective, qualitative metrics like preference polling (e.g., Likert scales), these may benefit from structured polling or checklists. Several initiatives have developed structured evaluation frameworks for LM models, integrating both quantitative benchmarks and human assessments. These checklists may be helpful to standardize evaluations, ensure comprehensive assessments, and identify potential risks \cite{gallifant2025tripod, sallam2024preliminary, chen2024stager}.

\begin{table}[t]
\centering
 \caption{Comparison of performance assessment strategies for generative AI in medical devices.}
    \label{tab:summary_of_methods}
    \vspace{1mm}
{\footnotesize
\begin{tabular}{>{\centering\arraybackslash}m{1.65cm}|p{2.95cm}|p{2.9cm|}p{2.9cm}}
    & \multicolumn{1}{>{\centering\arraybackslash}m{2.95cm}|}{\vspace{1mm}\textbf{Benchmark Evaluation}\vspace{1mm}} 
        & \multicolumn{1}{>{\centering\arraybackslash}m{2.9cm}|}{\vspace{1mm}\textbf{Human Evaluation}\vspace{1mm}} 
        & \multicolumn{1}{>{\centering\arraybackslash}m{2.9cm}}{\vspace{1mm}\textbf{Model-based Evaluation}\vspace{1mm}} 
        \\ \hline\noalign{\vskip .5pt}\hline
    
    \rowcolor{blue!5}
    \multicolumn{1}{>{\centering\arraybackslash}m{1.65cm}|}{\textbf{Concept}}
        & \multicolumn{1}{>{\raggedright\arraybackslash}m{2.95cm}|}{\vspace{1.5mm}Specific tasks using external datasets and predetermined metrics\vspace{1.5mm}}
        & \multicolumn{1}{>{\raggedright\arraybackslash}m{2.9cm}|}{\vspace{1.5mm}Use of expert annotations as the reference standard\vspace{1.5mm}}
        & \multicolumn{1}{>{\raggedright\arraybackslash}m{2.9cm}}{\vspace{1.5mm}Use of a model-based approach with human oversight\vspace{1.5mm}} 
        \\
        
    \multicolumn{1}{>{\centering\arraybackslash}m{1.65cm}|}{\textbf{Advantages}}
        & \multicolumn{1}{>{\raggedright\arraybackslash}m{2.95cm}|}
        {\vspace{1.5mm}\begin{itemize}[leftmargin=*,nosep,itemsep=1.5pt]
            \item Practical and available
            \item Head-to-head comparisons
            \item Scalable
            \end{itemize}\vspace{1.5mm}}
        & \multicolumn{1}{>{\raggedright\arraybackslash}m{2.9cm}|}
        {\vspace{1.5mm}\begin{itemize}[leftmargin=*,nosep,itemsep=1.5pt]
            \item Adaptable to new medical tasks
            \item Direct clinical relevancy
            \item Identification of model risks, biases, and errors
            \end{itemize}\vspace{1.5mm}}
        & \multicolumn{1}{>{\raggedright\arraybackslash}m{2.9cm}}
        {\vspace{1.5mm}\begin{itemize}[leftmargin=*,nosep,itemsep=1.5pt]
            \item Scalable
            \item Cost-effective
            \item Enables large-scale and real-time performance monitoring
            \end{itemize}\vspace{1.5mm}} 
            \\

    \rowcolor{blue!5}
    \multicolumn{1}{>{\centering\arraybackslash}m{1.65cm}|}{\textbf{Limitations}} 
        & \multicolumn{1}{>{\raggedright\arraybackslash}m{2.95cm}|}
        {\vspace{1.5mm}\begin{itemize}[leftmargin=*,nosep,itemsep=1.5pt]
            \item Limited in tasks and datasets
            \item Fail to capture real-world complexity
            \item Overfitting and Data leakage
            \end{itemize}\vspace{1.5mm}}
        & \multicolumn{1}{>{\raggedright\arraybackslash}m{2.9cm}|}
        {\vspace{1.5mm}\begin{itemize}[leftmargin=*,nosep,itemsep=1.5pt]
            \item Resource intensive
            \item Subjective and highly variable
            \item Prone to bias
            \end{itemize}\vspace{1.5mm}}
        & \multicolumn{1}{>{\raggedright\arraybackslash}m{2.9cm}}
        {\vspace{1.5mm}\begin{itemize}[leftmargin=*,nosep,itemsep=1.5pt]
            \item Burdensome validation
            \item Inter-model leakage
            \item Susceptible to adversarial attacks and hallucinations
            \end{itemize}\vspace{1.5mm}} 
            \\
\end{tabular}
}
\end{table}

\section{Model-based Evaluation}
Model-based evaluation, also referred to as a model as evaluator or MAE, relies on an independent model to evaluate another model. Though the principles of a model-based evaluation are not limited or unique to LM applications, the advancements in generative AI have expanded the effectiveness of model-based evaluations with the potential to support or replace human evaluation. 

\subsection{Description}
MAEs can be implemented in different LM applications including evaluation, alignment, retrieval, and reasoning tasks \cite{li2024generation}. MAEs have been applied to replicate human preference, determine the accuracy of results, detect hallucinations, and provide scores for metrics such as reliability, faithfulness, or robustness. The MAE mechanism can also be implemented into agentic implementations or reasoning tasks to improve the model's output \cite{gu2024survey, li2024llms}. Some examples of MAE applications include G-Eval \cite{liu2023g}, LLM Evaluation \cite{chiang2023can}, and a panel of LLM evaluators \cite{verga2024replacing}.

LLM-RadJudge uses a large language model (LLM) to evaluate errors within a radiology report \cite{wang2024llm}. Wang et al. were able to demonstrate that for identifying number of errors between a pair of radiology reports, a fine-tuned BioMistral-7B model achieved a Kendall's tau of 0.75 against a panel a 6 readers, which is within the variability of those readers on that dataset (0.74-0.82) \cite{wang2024llm}. VeriFact describes the use of an MAE within a retrieval implementation to assess brief hospital course summaries\cite{chung2025verifact}. At the sentence-level assessment for LLM-written summaries, VeriFact achieves 93\% agreement with the reference standard for a 3-label task of whether a sentence was "Supported," "Not Supported," and "Not Addressed" by facts. This agreement is higher compared to the 85\% agreement among individual readers, yet both VeriFact and humans have lower agreement rates for human-written summaries of 67\%. 

GREEN by Ostmeier et al. describes the use of an LLM to assess the factual correctness and errors in a radiology report achieving a Kendall's tau of 0.63, which is within the range of mean expert correlations (0.48-0.64) for fine-grained error counts\cite{ostmeier2024green}. In addition, Schmidt et al. assess the performance of LLMs to detect speech recognition errors in dictated radiology reports\cite{schmidt2024generative}. Studying a model's ability to detect internal inconsistencies, word omissions, nonsense phrases, and other error types within a report, they found that GPT-4 had an F1-score of 87\% and 94\% for clinically and not clinically significant errors, respectively. This was the best performing model among the four they tested in a dataset with strong inter-rater agreement.

\subsection{Advantages and Limitations}
Whether through the generation of the reference standard to allow head-to-head comparison of models for prespecified tasks or using an MAE to evaluate the performance of another LM model, these approaches reduce the burden on human annotators in testing studies. This results in lower study costs and improved efficiency. A model-based evaluator can also perform analyses at a scale not feasibly achieved by human evaluators. Moreover, these approaches may be useful to allow for evaluation of an implemented model, which can identify performance drift or bias more readily than if there were only pre-planned, intermittent retrospective analyses. These model-based evaluation studies would also be able to be performed at a faster rate than human-based evaluation studies, which may reduce any additional potential risk to patients. 

However, the evaluation standard for a model-based evaluator is high. If an MAE were to be implemented in the post-deployment setting, the model would have been previously demonstrated to adequately replicate the human assessment. This model would need intermittent continued performance evaluation to make sure it does not get impacted by performance drift, because any uncertainty or error in the MAE is propagated into the performance assessment of the model it is evaluating. This error propagation can lead to a performance misinterpretation of the model the MAE is evaluating. 

Just as LM models may have implicit biases from their training corpus, MAE implementations need to address and account for biases in position (e.g., position of example in a pairwise comparison affects performance), length (e.g., the MAE favors longer responses over concise ones), self-enhancement (e.g., the Evaluator prefers model outputs generated by the same model), and others\cite{li2024llms}. In addition, the MAE is also at risk of adversarial attacks that can misrepresent the evaluator's true performance on a benchmark dataset or mislead the model to be biased towards certain incorrect answers or hallucinations\cite{li2024llms}. 

\section{Summary}
The choice of evaluation method is guided by the model's intended use. For a tool-calling model that performs classifications, a vignette-styled benchmark dataset may offer insight into the model's final classification performance, but there would be a lack of assessment on any intermediate tool-calling steps. An additional dataset, human evaluation, or a well-validated MAE may further assess not only the model's performance on these missing steps, but by replicating the iterative information retrieval, the overall classification performance would better reflect the real-world setting. The human evaluation may also provide additional evaluation on the appropriateness of the model's intermediate steps or the final classification given the retrieved inputs; a level of detail not likely provided by a benchmark dataset or MAE. It is this wholistic consideration of a model's application that informs its evaluation.

We have identified three distinct strategies for LM evaluation, each presenting unique advantages and limitations (see summary in \autoref{tab:summary_of_methods}). Benchmark assessments enable the efficient testing of specific capabilities such as the accuracy of clinical reasoning, facilitating comparative head-to-head analysis. Nonetheless, they frequently fail to encapsulate the complexities of real-world scenarios and may become misleading if models adapt to optimize specific metrics rather than true clinical performance. 

Human evaluation offers the most clinically relevant assessment by leveraging domain expertise to evaluate outputs contextually. Human evaluators are capable of assessing nuanced aspects like clinical appropriateness and potential harm that automated methods might miss. The challenges associated with human evaluation include scalability, evaluator bias, subjectivity, and significant resource requirements. Additionally, inter-rater reliability can vary significantly, especially in intricate medical use cases. MAE techniques offer promising scalability and cost-effectiveness providing consistent application of evaluation criteria across large datasets at a lower cost. However, this approach has a high evaluation standard to minimize error propagation into the model it is evaluating. Current evidence suggests MAE methods may have a role in complementing rather than replacing human evaluation.

Several emerging strategies may enhance LM evaluation in clinical contexts. RLHF has shown promise in aligning LM outputs with human preferences and ethical considerations, potentially enhancing safety and reliability. A comprehensive evaluation strategy will likely benefit from combination approaches that integrate automated benchmarks, targeted human expert review, and model-assisted evaluation under human supervision. New evaluation methodologies and performance metrics may need to be developed to better quantify clinical reliability, safety, and potential risks. 

\section*{CRediT authorship contribution statement}
\textbf{Victor Garcia:} Conceptualization, Writing – review \& editing. \textbf{Mariia Sidulova:} Writing – original draft. \textbf{Aldo Badano:} Conceptualization, Writing – review \& editing.

\section*{Declaration of competing interest}
The authors have no conflicts of interest.

\section*{Disclaimer}
This article reflects the views of the authors and does not represent the views or policy of the U.S. Food and Drug Administration, the Department of Health and Human Services, or the U.S. Government.  The mention of commercial products, their sources, or their use in connection with material reported herein is not to be construed as either an actual or implied endorsement of such products by the Department of Health and Human Services.

\bibliographystyle{unsrtnat}   
\bibliography{bibliography_revision_1}

\end{document}